\documentclass[conference]{IEEEtran}
\usepackage{cite}

\ifCLASSINFOpdf
   \usepackage[pdftex]{graphicx}

\else

\fi

\usepackage{array}
\usepackage{url}
\begin{document}

\title{Bias and Fairness in Computer Vision Applications of the Criminal Justice System}

\author{\IEEEauthorblockN{Sophie Noiret} 
\IEEEauthorblockA{Vienna University of Technology\\
Computer Vision Lab\\
Vienna, Austria\\
Email: snoiret@cvl.tuwien.ac.at} 
\and 
 \IEEEauthorblockN{Jennifer Lumetzberger }
\IEEEauthorblockA{Vienna University of Technology\\
Computer Vision Lab\\
Vienna, Austria\\
Email: jennifer.lumetzberger@tuwien.ac.at}
\and 
\IEEEauthorblockN{Martin Kampel} 
\IEEEauthorblockA{Vienna University of Technology\\
Computer Vision Lab\\
Vienna, Austria\\
Email: martin.kampel@tuwien.ac.at}
}

\maketitle

\begin{abstract}
Discriminatory practices involving AI-driven police work have been the subject of much controversies in the past few years, with algorithms such as COMPAS, PredPol and ShotSpotter being accused of unfairly impacting minority groups. At the same time, the issues of fairness in machine learning, and in particular in computer vision, have been the subject of a growing number of academic works. In this paper, we examine how these area intersect. We provide information on how these practices have come to exist and the difficulties in alleviating them. We then examine three applications currently in development to understand what risks they pose to fairness and how those risks can be mitigated.
\end{abstract}

\ifCLASSOPTIONpeerreview
\begin{center} \bfseries EDICS Category: 3-BBND \end{center}
\fi

\IEEEpeerreviewmaketitle

\section{Introduction}
In 2016, journalists from ProPublica put the spotlight on the discriminatory practices in AI-driven predictive policing through their analysis of the Correctional Offender Management Profiling for Alternative Sanctions tool, or COMPAS\footnote{J. Angwin, J. Larson, S. Mattu and L. Kirchner "Machine Bias" \textit{ProPublica} (May 23rd, 2016).Accessed on: October 15, 2021. \url{https://www.propublica.org/article/machine-bias-risk-assessments-in-criminal-sentencing}}. While Northpointe (which renamed itself Equivant in 2017), the company responsible for the development of this tool, disputes these claims, they nevertheless highlight the dire consequences of automated bias (in this case racial bias) in the justice system. But the use and possible misuse of AI-driven tools by the police and the court system has not stopped. In an article published by Vice in July 2021\footnote{T. Feather "Police Are Telling ShotSpotter to Alter Evidence From Gunshot-Detecting AI" \textit{Vice} (July 26th, 2021). Accessed on: October 15, 2021.  \url{https://www.vice.com/en/article/qj8xbq/police-are-telling-shotspotter-to-alter-evidence-from-gunshot-detecting-ai}}, the journalists allege that ShotSpotter, a system used by the Chicago police that detects the sound of gunshots and triangulate their positions, has its output regularly modified by request of the police, in addition to targeting minorities and not performing to the levels of accuracy promised by the company. Once again, ShotSpotter  denies these claims. 

Computer vision is another area of AI plagued by bias. The popular dataset ImageNet, for instance, exhibits a lack of diversity and problematic labelling of images \cite{yang_towards_2020}. In their seminal work, J. Buolamwini and T. Gebru show the differences in errors rates for gender classifiers across racial groups \cite{Buolamwini.2018}. Addressing this issues is of paramount importance, so that they don't overshadow the genuine good that can come from both of these areas: recent advances in computer vision allow Neural Network to detect brain cancer with  98.67\% accuracy \cite{siar_brain_2019} while surveillance drones in national park can automatically detect poachers \cite{bondi_spot_nodate}.

These issues are being identified by policy makers, as several legislations addressing them exist across the world at different stages of implementation. The GDPR\footnote{Regulation (EU) 2016/679, \textit{Official Journal of the European Union}. Accessed on: October 15, 2021. \url{https://gdpr-info.eu/}}, which came into effect in May 2018, introduces concepts of both fairness and transparency \cite{malgieri_concept_2020}. The proposed AI Act\footnote{Proposal for Artificial Intelligence Act (COM(2021)206).  Accessed on: October 15, 2021. \url{https://eur-lex.europa.eu/legal-content/EN/TXT/?uri=CELEX\%3A52021PC0206}} takes a risk-based approach to AI regulation, with high-risks applications being obligated to provide technical documentation including "information about the validation and testing data [..]; metrics used to measure accuracy, robustness [...] as well as potentially discriminatory impacts". Amongst high-risks applications are "Ai systems intended to be used by law enforcement authorities [...] to assess the risk of a natural person for offending or re-offending[...] as polygraphs and similar tools[...] for evaluation of the reliability of evidence [...]predicting the occurrence or reoccurrence of an actual or potential crime". The AI Act would also ban the use of real-time biometric data except in specific cases. In the US, the  Facial Recognition and Biometric Technology Moratorium Act\footnote{116th Congress (2019-2020): Facial Recognition and Biometric Technology Moratorium Act of 2020.  Accessed on: October 15, 2021.   \url{https://www.congress.gov/bill/116th-congress/senate-bill/4084} }, which would temporarily ban the use of face recognition, was re-introduced on June 25th 2021.
While these legislations are not yet adopted or enacted, others voluntary certifications exist, such as Malta's certification program\footnote{Malta.AI Taskforce, "Malta: Towards Trustworthy AI", August 2019.  Accessed on: October 15, 2021. \url{https://malta.ai/} }. Applicant to this certification must follow the guidelines provided by Malta AI taskforce, which focus on human autonomy, prevention of harm, fairness and explicability. 

This paper examines the actual use of algorithmic fairness in AI-driven computer vision applications in the justice system. We do so through semi-structured interviews with developers involved in research projects for criminal justice. The rest of the paper is structured as follows. Chapter 2 lays a terminological basis for the subsequent chapters by explaining relevant terms. Chapter 3 highlights both the need for and difficulties in introducing fairness in AI-driven computer vision application in the justice system. Chapter 4 presents 3 applications being developed in this area, focusing on technical specification, discrimination risks and recommendations. Chapter 5 concludes this paper.

\section{Terminology}

Before we can proceed to how fairness can be introduced to AI and ML systems, we need to start by considering the key-terms that we - and many others - use. Particularly terms stemming from philosophy and law like bias, discrimination and fairness require a common understanding, as they represent core elements we want to address in our paper. As the choice of metric for fairness can also change whether or not a system is considered fair, we need to present some of the metrics that come into play in the rest of the paper. 

\subsection{Bias, Discrimination and Fairness}

We start by defining the terms of bias and discrimination in conjunction, as they are closely related. 
A rare and thorough analysis of what bias means in the context of algorithms and automated decision making was provided by Friedman and Nissenbaum \cite{Friedman.1996}, who consider algorithms as biased, in that they have the potential to "systematically and unfairly discriminate against certain individuals or groups of individuals in favor of others". The authors furthermore explain what "unfairly discriminates" means in the context of computer systems and algorithms, namely "if it denies an opportunity or a good or if it assigns an undesirable outcome to an individual or group of individuals on grounds that are unreasonable or inappropriate". \\
This definition is insofar useful, as it encompasses both the terms of bias and of discrimination and shows how discrimination should be considered as the biased action, the act of treating someone unfairly, bridging to the term of fairness.
\subsubsection{3 types of bias} 

Having seen that an algorithmic system is biased if it "unfairly discriminates against certain individuals or groups", it is worth considering the origin(s) of the bias. 
Generally, there are three different types of algorithmic biases that can be differentiated: (1) a pre-existing bias; (2) a technical bias; (3) an emergent bias \cite{Friedman.1996}.

 A pre-existing bias is a bias that is prevalent within our societies, and is the preconceived, internal opinion about individuals or groups. A pre-existing bias stems from the social institutions, from our norms and attitudes and generally manifests itself as discriminative through our practices. This also means that the pre-existing bias can reflect solely the personal opinion and prejudice of one individual - in this case of someone who is responsible for the design of the algorithmic system. But the bias can also be systemic within our society and thus way more difficult to address. 
 In many examples that we will discuss here, pre-existing biases are introduced into automated decision making and machine learning algorithms implicitly and unconsciously, encoded via disparate impact \cite{Feldman.2015}. 
The second type of algorithmic bias is inherent of the technology itself. Technical biases usually emerge within the design process of the algorithms and are the results of limitations of computer tools such as hard- and software. They can be the result of errors in coding, and in the construction and design of the algorithm \cite{Kitchin.2016}. But they also can emerge when system designers attempt to digitalise human qualities, when they are trying to make fundamental human aspects machine readable. 

The third type of algorithmic biases are emergent biases. These do not exist (per se) in the technology straight “out of the box.” Instead, they emerge over time as new knowledge is created that can't be integrated into the algorithm. Or through the interaction between the technology and the users, producing outcomes that were not intended or considered by the system designers. 
As automated decision making technologies are increasingly integrated in everyday societal practices, individuals have to adapt to incorporate these technologies into their routines. In most cases, this means that they need to make themselves "machine-readable" \cite{Kreissl.2014}. If such technologies are applied in areas with a low technological knowledge, amongst individuals which have difficulties reading, hearing, seeing, etc., biases quickly emerge.
        \\ 
\begin{figure*}[h]
  \centering
  \includegraphics[width=\linewidth]{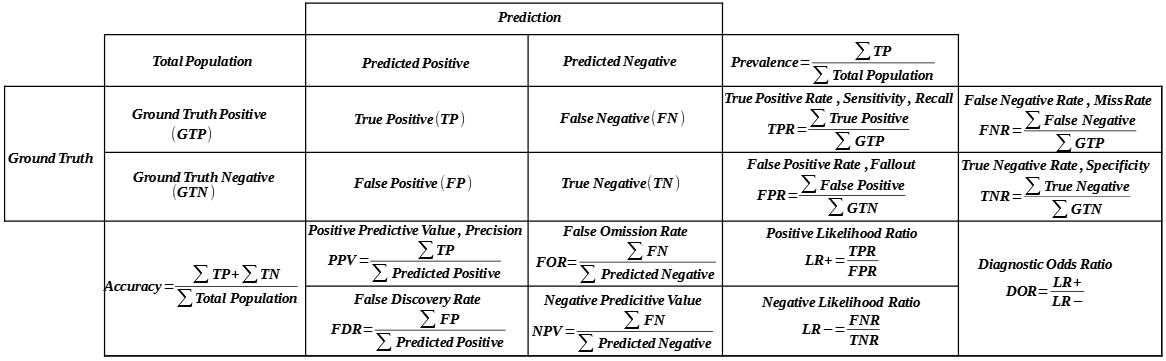}
  \caption{Confusion Matrix and Associated Metrics }

  \label{ConfusionMatrix}
\end{figure*}
\subsubsection{Different types of discrimination} 
As with bias, there are also different aspects of discrimination, that need to be considered. Although the concept of discrimination as an unfair treatment of individuals or groups remains the same, this treatment can be based on different attributes. Direct discrimination describes the unfair treatment that is based on protected grounds, such as age, gender, ethnicity, disability, creed, sexual orientation, etc. However, unfair treatment can also be the outcome of indirect discrimination, through proxies\cite{Mann.2019}. It is worth noting, for instance, that COMPAS allegedly does not explicitly use race as an input. However, the 173-questions long questionnaire includes sections about the neighbourhood and family history of the defendant. \\ 
Besides direct and indirect discrimination, there are also two other forms of discrimination that need to be mentioned here, as they increasingly emerge through the use of algorithms and automated decision making. These are intersectional discrimination and emergent discrimination. \\
Intersectional discrimination addresses the more complex situations of discrimination that occur through a combination of discriminating characteristics. The idea behind intersectional discrimination is that, for example, the combined discrimination based on gender and ethnicity for women of colour is experienced differently than single entities of discrimination based on ethnicity and based on gender \cite{Mann.2019}. Particularly through algorithmic profiling and individualised and personalised decision-making, these intersection of discriminating identities risk to occur more often, because much more individual characteristics are used as a method to assess individuals. Which also brings us to the aspect of emergent discrimination. As with emergent bias seen before, also discrimination can occur over time, without having accounted for the potential of future discrimination situations. This is even more the case when intersectional discrimination is taken into account, where discrimination might emerge due to a combination of potentially discriminating characteristics. \\ 

\subsection{Metrics}
An appropriate choice of a metric can change our vision of the fairness of an algorithm. Fairness metrics are often separated into group fairness, which looks at the outcome of the algorithm for different groups, and individual fairness, which is achieve is similar individual are treated similarly \cite{Caton2020FairnessIM}. In cases of proprietary algorithms, individual fairness can be difficult to measure: without access to the system, one cannot, for instance, determine what the output would have been if an individual had not been a part of a protected group. As our aim here is to give an overview of the metrics relevant to our study, we will focus on group fairness and particularly on Confusion-matrix based metrics. Figure 1. presents a confusion matrix and the associated terminology.

\underline{Demographic Parity}: The probability of being predicted positive is equal across groups. Closely related to this notion is Disparate Impact \cite{Feldman.2015}, which relaxes strict Demographic Parity by only demanding that the ration between the probability of being predicted positive across groups is under a certain threshold. 
\underline{Conditional Demographic Parity} \cite{corbett-davies_algorithmic_2017-1}: Also called conditional non-discrimination \cite{kamiran_quantifying_2013}. Controlling for a set of legitimate factors, the probability of being predicted positive is equal across groups.

\underline{Equal opportunity}: True Positive Rate is equal across groups.

\underline{Equalize Odds}: True Positive Rate and False Positive Rate are equal across groups.

\underline{Conditional use accuracy}: Negative Predictive Value and Positive Predictive Value are equal across groups. When the result is not a binary classification but a risks-core, this metrics becomes "calibration": given a particular score, the probability of being ground-truth positive (resp. negative) is equal across groups.  

All metrics cannot be achieved at the same time. Particularly, Equalized Odds and Conditional use accuracy can only be achieved at the same time if the prevalence is equal across group, or in the case of a perfect classifier \cite{berk_fairness_2021}. 

\section{Discrimination in Computer Vision, Forensics and Predictive Policing}
Discrimination and biases in computer vision, as well as in policing, have been the focus of a large number of work.
\subsection{Bias in Computer Vision}
D Klare et al. \cite{Klare.2012} analyse six different face recognition algorithms and find that the commercial as well as the non-trainable algorithms have lower matching accuracies on the same cohorts (females, Blacks, and age group 18-30) than on other cohorts. Two facial analysis benchmarks, the datasets IJB-A and Adience, consist of 79.6-86.2\% of people with lighter skin, leading to higher performance for this group \cite{Buolamwini.2018}. Additionally, gender classification systems from companies with large investments in artificial intelligence and facial analysis technologies, namely Microsoft, IBM and Face++, are analysed. Darker-skinned females have the highest error rates (up to 34.7\%), while the maximum error rate for white men is 0.8\% \cite{Buolamwini.2018}. Even, in wide-scale and largely-used datasets such as ImageNet, biases can arise from inappropriate labelling of images, the use of non-visual concept or the lack of image diversity \cite{yang_towards_2020}. 

This is not a purely academic matter, as despite these concerns, the global facial recognition market is expected to grow at a compound annual growth rate of 14.5\% from 2020 to 2027\footnote{"Global Facial Recognition Market Size, Share \& Trends 2020-2027", \textit{Business Wire}, April 2020 \textit{Business Wire}. Accessed on: October 15, 2021. \url{https://www.businesswire.com/news/home/20200409005340/en/Global-Facial-Recognition-Market-Size-Share-Trends-2020-2027}}. The consequences could be dire: in 2018, the American Civil Liberties Union (ACLU) conducted a study on Amazon's Rekognition tool, in which they built a database using publicly available arrest photos and searched for matches amongst the member of congress. 28 of them were falsely identified, with people of color accounting for 39\% of the false matches, while making up only 20\% of member of Congress. 

On May 18, 2021, Amazon has announced an extension of their moratorium on police use of their software\footnote{J. Dastin " Amazon extends moratorium on police use of facial recognition software", \textit{Reuters}, May 18th, 2021. Accessed on: October 15, 2021.  \url{https://www.reuters.com/technology/exclusive-amazon-extends-moratorium-police-use-facial-recognition-software-2021-05-18/}}, which may be wise considering how bias in CV could have a compounding effect with the existing bias in policing. 
\begin{table}[!t]
 \caption{COMPAS Performance on Confusion Matrix Metrics}
 \label{table_example}
 \centering

 \begin{tabular}{|c||c|c|c|}
 \hline
   & All Defendants & Black Defendants & White Defendants\\
 \hline
 FP rate &  32.35 & 44.85 & 23.45 \\
 \hline
  FN rate &  37.4 & 27.99 & 47.72 \\
   \hline
  PPV &  0.61 & 0.63 & 0.59 \\
   \hline
NPV &  0.69 & 0.65 & 0.71 \\
\hline
\end{tabular}
\end{table}
\subsection{Policing}
While computer vision only started being developed in the 1960s, policing has long history fraught with bias and discrimination \cite{goff_racial_2012}. Even in cases of not explicitly discriminatory laws, the targeting and over-policing of minority groups has led them to be over-represented in every aspect of the criminal justice system \cite{browning_stop_2021}. The practice of the now debunked pseudo-science of phrenology in criminal cases in the 19th century shows that the "scientific" police was not exempt from this issue \cite{rafter_murderous_2005}. This pre-existing bias can manifest itself in the result of policing algorithms.      

\subsubsection{Predictive Policing} ProPublica analysis of COMPAS shows that while the algorithm's accuracy is comparable for white and black and defendants (62.5\% and 62.3 \% respectively), the nature of the errors leads to black defendants being more likely to be misclassified as high-risk (False Positive Rate) while white defendant are more likely to be misclassified as low-risk (False Negative Rate). Table 1 sums up the finding of ProPublica. This illustrates the problem of the incompatibility of metrics that we introduced earlier: while COMPAS does not satisfy Equal Opportunity or Equalize Odds criteria, it performs rather well with regards to Conditional Use Accuracy. This is the basis of Northpointe's defense in their answer to ProPublica\footnote{Equivant, "Response to ProPublica: Demonstrating accuracy equity and predictive parity", Decembre 2018. Accessed on: October 15, 2021. \url{https://www.equivant.com/response-to-propublica-demonstrating-accuracy-equity-and-predictive-parity/}  }. Unless the recidivism rate is equal across group, a choice has to be made between those metrics. 

It is unclear whether or not disparate impact liability applies in case of predictive policing: While Tiwari \cite{tiwari_disparate-impact_2019} argues that the Safe Streets Act imposes disparate impact liability on police departments, Selbst points out in \cite{selbst_disparate_2018}  that disagreements about the meaning of discrimination would prevent the application of disparate impact in the case of predictive policing. They also point out in \cite{barocas_big_2016} that "data mining can reproduce existing patterns of discrimination, inherit the prejudice of prior decision makers, or simply reflect the widespread biases that persist in society."

This echoes the findings of \cite{browning_stop_2021} that police and crime data are biased along racial and ethnic lines and do not reflect empirical reality. This can create a feedbackloop, in which biased data are used to justify the over-policing of minority group, leading to more biased data. If no precautions are taken, predictive policing can exacerbate this problem: Lum and Isaac find that PredPol, another predictive policing algorithm, is more likely to target low-income, Black communities \cite{lum_predict_2016}. Even more recently, journalists from Motherboard found that the sensor for ShotSpotter were placed almost exclusively in neighborhoods of Chicago inhabited by Black communities.
This is not a purely American problem: the European Union Agency for Fundamental Rights published a survey showing that that police stops more often concern men, young people, as well as people who self-identify as belonging to an ethnic minority, who are Muslim, or who are not heterosexual\footnote{European Union Agency for Fundamental Rights (FRA) (2021), "Your rights matter: Police stops, Fundamental  
Rights Survey". Accessed on: October 15, 2021. \url{https://fra.europa.eu/en/publication/2021/fundamental-rights-survey-police-stops}}.

\subsubsection{Forensics}
This over-targeting of minority groups also appears in the databanks used for forensics science \cite{noauthor_racial_nodate}: in the US, although Black people make up only 13.26\% of the U.S. population, they make up 34.47\% of the DNA database. Once again, we see a feedback loop at play, as crimes committed by a member of a group underrepresented in the DNA databank are less likely to be solved. Additionally, the police may be inclined to spend an greater amount of their limited time and resources on cases where they have found a DNA match \cite{DNA}.

This existing bias can compound with algorithm appreciation, an emergent bias in which people display an over-reliance on algorithmic judgment \cite{logg_algorithm_2019}. This is exacerbated by claims made by some software companies: in response to the article published by VICE, ShotSpotter has issued a statement claiming hat "ShotSpotter forensic evidence is 100\% reliable and based entirely on the facts and science". While they report a 97\% rate accuracy, confirmed by the consulting group Edgeworth Analytics\footnote{Edgeworth Analytics, "Independent Audit of the ShotSpotter 
Accuracy", July 22, 2021. Accessed on: October 15, 2021. \url{https://-edgeworthanalytics.com/independent-audit-of-the-shotspotter-accuracy/}}, the MacArthur Justice Center found that 89\% of ShotSpotter deployments in Chicago turned up no gun-related crime\footnote{ MacArthur Justice Center, "ShotSpotter creates thousands of dead-end police deployments that find no evidence of actual gunfire", May 03, 2021. Accessed on: October 15, 2021. \url{https://endpolicesurveillance.com/}}. 

These conflicting reports speak to the lack of transparency in those algorithms: citing trade secret, neither Northpointe nor ShotSpotter have made their algorithms publicly available, making them effectively black-boxes. 

\section{Applications}

In order to realize their full potential in crime prevention, crime solving and safety improvement, fairness and bias must be taken into account when developing and using computer vision applications. Thus, we have conducted semi-structured interviews with developers and PhD students from Vienna University of Technology and  involved in such projects between April and May 2021.

\subsection{Methodology}
The interviews were conducted with one or two developers from each project. The developers were asked to briefly describe the project, after which the interviewer asked follow-up questions if necessary in order to cover the following topics:\\
\underline{Finality}:
\begin{itemize}
    \item What is the aim of the project ?
    \item Who is it going to be used by ? Who is it going to be used on ?
\end{itemize}

\underline{Description of ML strategy}:
\begin{itemize}
    \item What is the output of the algorithm? 
    \item What category of ML is used ?
    \item How did you select the model ? 
    \item Will there be human intervention ?
\end{itemize}

\underline {Data}:
\begin{itemize}
    \item What are the training data ? Are synthetic data used? 
    \item How were they acquired ? How much of it is there ? 
    \item Is it representative of the population it is going to be used on ? (How do you know ?)
    \item Is there pre-treatment in order to detect/mitigate bias ?
\end{itemize}

\underline{Explainability/ Interpretability}:
\begin{itemize}
    \item Is there any interpretability technique ?
    \item Are the decisions and their explanations logged ?
\end{itemize}

\underline{Performance}:
\begin{itemize}
    \item Which metrics are used to measure performance ?
\end{itemize}

\underline{Fairness}:
\begin{itemize}
    \item Were fairness and transparency considered ?
    \item Do you foresee any possible issues regarding fairness and discrimination ? If so, what are they ?
\end{itemize}
Follow-up questions regarding type of fairness, metrics and mitigation techniques were initially considered, but as fairness was not integrated into the projects, they became irrelevant.

\begin{table*}[!t]
\caption{Summary of findings}
\label{table_example}
\centering
\begin{tabular}{|p{1.7cm}||p{4.8cm}|p{4.5cm}|p{4.5cm}|}
\hline
\textbf{Application} & \textbf{Footwear Impressions} & \textbf{ Handwriting Comparisons} & \textbf{Prison Surveillance} \\
\hline
 \textbf{Finality} & To be used by forensics experts on suspects &  To be used by forensics Experts on suspects  & To be used by prison personnel on inmates \\
\hline
 \textbf{ML Strategy} & Triplet of CNN & Local Naive Bayes Nearest-Neighbour & YOLOv3 CNN \\
\hline
 \textbf{Training Data} & 4000 images of 300 pairs of shoes on differents materials & 2500 handwrtiting samples from prison population &85000 frames over 20 indoor sequences recorded at the university \\
\hline
 \textbf{Fairness Risks} & Effects of representativity and lack thereof \newline Effects of gender on footwear impressions  &  Effects of native scripts, gender, left-handedness \newline Feedback loops & Effects of different body types and mobility on person detection \newline  Effects of cultural background on aggression detection\\

\hline
\end{tabular}

\end{table*}

\subsection{Search and comparison of footwear impressions} 

Given an image of a footwear impression, this first project creates a list of best possible matches from a dataset. This helps in identifying the model and brand of footwear, or in knowing whether two impressions were made by the same shoe. This process, usually done by human experts, can be time-consuming when the experts have to sift through a large database of known footwear impressions.

The chosen approach to create this list of similarities is a TripNet neural network based on the architecture described in \cite{TripNet}. This method aims at learning local feature descriptors to learn similarities. This is achieved by using three parallels CNN with shared weights. During training, three input samples are forwarded through the branches. Two of these input form a positive pair, meaning that they come from the same class and one is from a different class (thus forming negative pairs with the other two). The loss function then penalizes positive pairs with great distance and negative pairs with short distance and the error is back-propagated. Since the weights are the same in all three branches, only one of them is used in testing and use. The performance are measured through mean average precision, recall and cumulative match score. \\
In order to train the algorithm, a dataset is created using footwear impression on a variety of materials. The dataset and the method of acquisition are described in \cite{ImpressDataset}. While the creation of this dataset certainly allows for better machine learning training, it poses the question of representation bias (lack of diversity in the dataset) versus historical bias. According to \cite{StudyHomicide}, 92\% of those convicted of homicide in Europe are men. Should a dataset intended to be used in policing reflect this statistic, or should it aim at a more balanced representation ? It is in fact unclear how gender differences translate into footwear impressions: the difference could be in model of footwear, or depth, shape and size of impression. In this case, it would be useful to know if a dataset that does not represent the real world results in degraded performance for one group. Ultimately, the decision should be made taking into account the potential harm that could come from it, even if the decision may be at odds with the available data.

\begin{figure}[h]
  \centering
  \includegraphics[width=\linewidth]{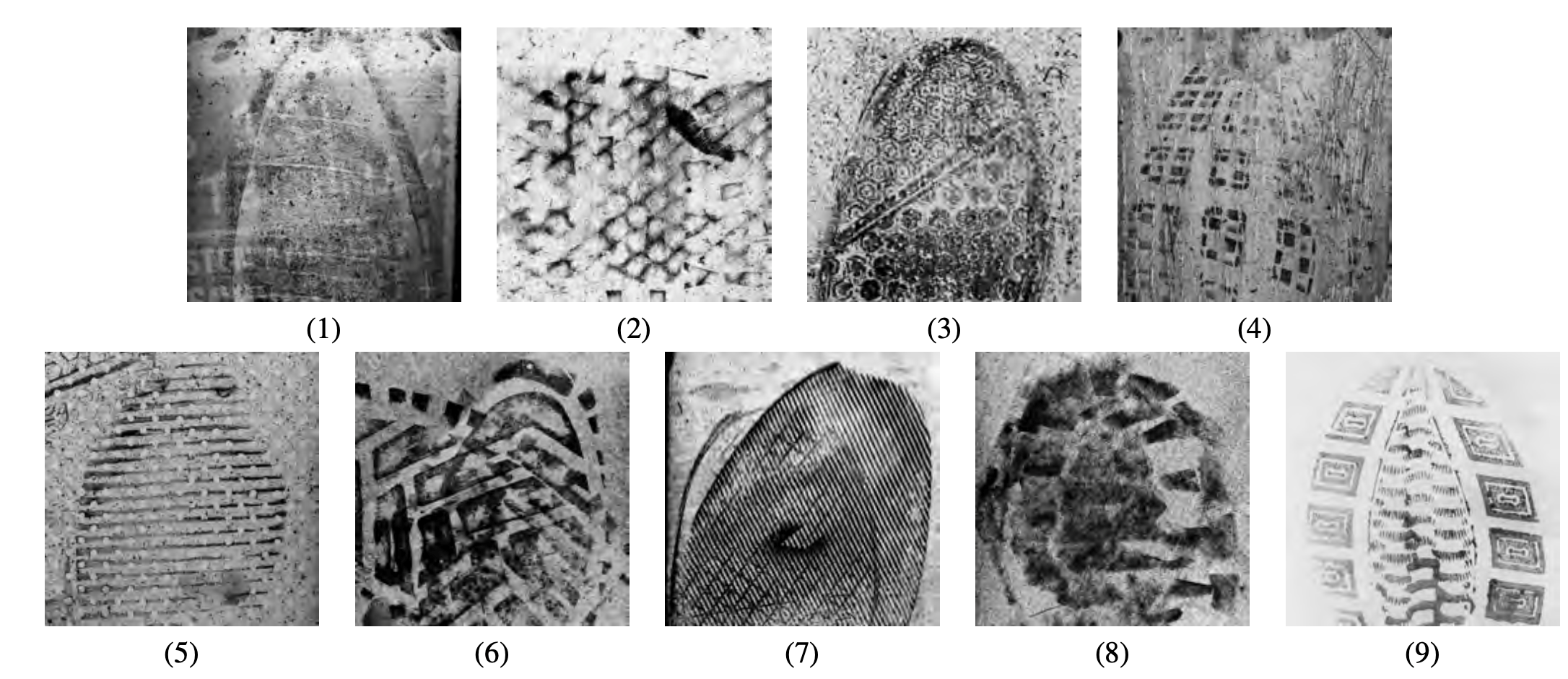}
  \caption{Realistic impression from the "Impress" dataset, taken from \cite{Keglevic2018LearningFF}}
  \label{ImpressDataset-pic}
\end{figure}
\subsection{Search and comparison of handwritings}
There is a strong demand in the sector of forensic analysis for algorithms to facilitate the work of experts, as this second project in this sector highlights. The aim is to create a list of best possible matches for a handwritten sample. This tool is by forensics experts in the Austrian police.

The approach chosen here is similar to the one described in \cite{mohammed_normalised_2017}, in which a  Local Naïve Bayes Nearest-Neighbour algorithm is used to assign an image, i.e. a handwritten sample, to a writer, i.e. a class. The images are first converted to grayscale, and the SIFT algorithm is then used to detect keypoints and store them and their descriptors in vectorial form. In a classic Naïve Bayes Nearest-Neighbour classifier as described, the algorithm finds the distance to the nearest neighbor of the descriptor in each class. The process is repeated for all the descriptors in the image to classify, with the distances being added in each class. The classes are then ranked based on the total distance, the class lowest total distance being the first. The Local Naïve Bayes Nearest-Neighbour Classifier improves the speed and accuracy of this process by only looking at classes that are close to the considered descriptor.
Additionally, a visual explanation for the ranking is provided by placing side-to-side keypoints in the original image and the corresponding nearest neighbor in the ranked document. 

This application is being developed using the mandatory writing samples given by the prison population in Austria. Each sample consist of a pre-determined text in german which has to be written twice, once in cursive and once in block letters. Each writer contributes at least two sample to the dataset. The first risk with this dataset is that it mirrors some of the issues we've seen before in DNA banks: crime committed by groups that are already under-represented in the dataset are less likely to be solved, creating a feedback loop. This potential feedback loop risks affecting predominantly foreigners, as according to the Institute for Crime \& Justice Policy Research, foreigners in Austria are over-represented in the prison population, making up 53.2\% of the prison population but only 16.7\% of the population\footnote{Institute for Crime \& Justice Policy Research, "World prison Brief". Accessed on: October 15, 2021. \url{https://www.prisonstudies.org/country/austria} }. However, in contrast with DNA sample in the US, handwritten samples are collected from people who have been convicted, not just arrested.
While it could be argued that one's nationality should have no influence on their handwriting, other factors could come into play, such as left-handedness or whether the Latin alphabet is the primary script used by the writer. 

An unbalanced dataset can also lead to adverse effect for the minority group: for instance, it is unknown if the algorithm might is able to distinguish between two left-handed writers. The study of the balance and representativity of the dataset is however hindered by the fact that, for privacy reasons, the developers do not have access to the file of the person who has given the sample. 
\subsection{Machine Learning of motion patterns in the penal system}

\begin{figure}[h]
  \centering
  \includegraphics[width=\linewidth]{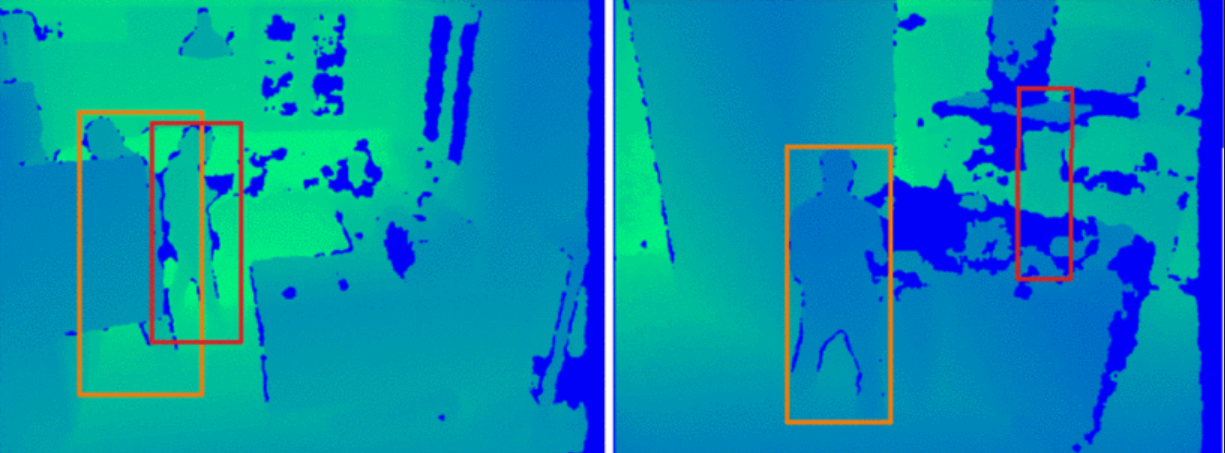}
  \caption{3D image, taken from \cite{9412979}.}
  \label{aggression-pic}
\end{figure}

This application focuses on enhancing the security of prison inmates and employees by detecting potentially dangerous behavior in 3D sensor data in a fully automated way. This is accomplished via real-time detection of people and a temporal analysis of their movement patterns as well as body poses based on learning techniques. On this basis, the developed system can alert the security personnel of potential dangers such that intervention can take place at an early stage. 
Utilizing 3D data enables unobtrusive monitoring, preserves the privacy of the people involved, and avoids data protection issues that have to be taken into account when utilizing video cameras.
An example image of the processed data is illustrated in Figure \ref{aggression-pic}. When predefined behavior models match recorded motion patterns, an alert is raised.

The first step in this application is person detection. This is accomplished by a CNN based on a modified YOLOv3 architecture \cite{yolo}. YOLO (You Only Look Once) is a fast single-staged CNN which predicts bounding boxes and class probabilities. It is modified here in order to only output the 'person' class and trained from scratch on privacy-preserving 3D data. This architecture is particularly well-suited for embedded solution, as it is fast and computationally efficient.

The use of 3D data instead of RGB can help solve a persistent problem in person detection and recognition: the difference in performance based on skin tone. 
But this alone does not ensure fairness: In \cite{banerjee2021reading}, the authors show that medical AI systems are able to predict the race of the patient in medical images such as x-rays and CT-scans, even when controlling for possible proxies such as body mass index, tissue density, disease label or age and sex. Since these medical images, like depth data, do not show skin tone, it cannot be assumed that depth images occludes this information. 
However, factors such as height, weight or mobility could lead to degraded performance if the dataset lacks in diversity. 

Regarding behaviour analysis, a core problem lies in defining and classifying what aggressive motion patterns are, how they manifest themselves and how they differ from non-aggressive motions. These questions need to be covered, answered and documented, as it also contributes to the transparency and explainability of the system. This also assists to consider, whether defined aggressive motion patterns look the same for each individual, or whether differences might appear on the basis of gender, age, body type, or disabilities. If one inmate triggers more frequently an alarm than another based on different (non-aggressive) motion patterns that are not covered by the initial training data set, this might lead to unfair treatment.
\subsection{Discussion}
The following risks and impediments regarding fairness span across all three of this applications.   
\subsubsection{Considering fairness}
Despite the sensitive nature of all the application investigated, fairness has not considered in any of the project at that stage. However, when prompted, developers expressed interest in the subject and volunteered their opinions on possibles biases. It was also pointed out that no application is intended to be used in a fully automated system (there is always human intervention). The issue of algorithmic fairness, while it is gaining traction in the machine-learning community, has not yet reached a point where its inclusion is par for the course, even in sensitive application. This does not appear to be result of deliberate resistance or unwillingness, but rather to a lack of exposure to the concept. The first step improving fairness in real-life application must be to improve awareness of AI-developers about the subject. However, there are significant roadblocks to implementing fairness in the projects considered.

\subsubsection{Choosing and accessing protected attributes}
Developers cited the lack of knowledge about which attribute to consider for fairness as a barrier to implementing it. Moreover, in systems with several stages, such as the Machine Learning of Motion Patterns in the Penal System, relevant attributes can differ across stages: height, weight or mobility can degrade the performances in person detection, while neurodivergence or cultural norms can have an influence in behavior recognition. 
In those cases, it is yet unclear if fairness should be considered and addressed in each stage or for the system as a whole.

When the protected attributes are either not accessible or not known, more advanced de-biasing technique can be used: Fair Class balancing \cite{yan_fair_2020} can modify the dataset by automatically re-sampling without Observing Sensitive Attributes. It is also possible to use distributionally robust optimization \cite{hashimoto_fairness_2018} or Adversarially Reweighted Learning \cite{sattigeri_fairness_2019} to improve the fairness of the algorithm.

It must also be noted that protected attributes such as race, gender or religion are social constructs that are culturally defined and cannot be directly translated from one country to another. To take race as an example, in the 2010 US census\footnote{U.S. Census Bureau, "U.S. Census Bureau QuickFacts: United States. Accessed on: October 15, 2021. \url{https://www.census.gov/quickfacts/fact/table/US/PST045219}}, people could self-identify as "White", "Black or African American", "American Indian and Alaska Native", "Asian", or "Native Hawaiian and Other Pacific Islander".  
In contrast, in the survey by the European Union Agency for Fundamental Rights, the target groups were "Immigrants and descendants of immigrants from [place of origine]" (with place of origin amongst Turkey, North Africa, Sub-Saharan Africa and  South Asia and Asia), "Recent immigrants from other non-EU/EFTA countries", "Russian minorities" and "Roma".
Moreover, access to personal data, even to ensure fairness, will be difficult before the the AI Act is adopted and enacted. Article 9 of the GDPR prohibits the use of personal data, with a few exceptions. As pointed out in \cite{veale2021demystifying}, the proposed AI Act would provide an exception once it is implemented, if it is "strictly necessary for the purpose of ensuring bias monitoring, detection and correction".
\subsubsection{Choosing fairness metrics}
The open question of which definition of fairness and which metric to use is another impediment. Applications destined for commercial use especially need to comply with regulations to access the market. While the proposed AI act imposes conformity assessment with regard to the quality of the data and the accuracy, robustness and security of the system, it does not give concrete metrics to use. According to  \cite{veale2021demystifying}, establishing precise standards is the responsibility of the European Committee for Standardisation. In the accompanying impact assessment, the European Commission estimates that these standards will be available within 3 to 4 years.
In the meantime, practitioners need to make sure that the regulations they follow are coherent with the country the application is destined to be used in: while investigating the claim of bias mitigation in automated hiring systems in the UK, the authors of \cite{10.1145/3351095.3372849} note that 2 out of the 3 systems investigated explicitly aim at complying with US Equality law as implemented with the 4/5th rule, which is not applicable in the UK. However, concerning EU laws, the authors of \cite{Wachter2021BiasPI} and \cite{Wachter2020} argue that Conditional Demographic Disparity is "the most compatible with the concepts of equality and illegal disparity as developed by the European Court of Justice". As we've seen, this concept is closely related to demographic parity, the basis for disparate impact.

\section{Conclusion}

In this work, we have presented three applications which stand at the intersection of areas particularly at risk of discrimination: computer vision and policing.     

We first laid a terminological basis in order to clarify technical, sociological and legal terms before presenting a short overview of known risks and instance of discrimination within the areas of interest.
Finally, the applications are described from a technical standpoint and we outline associated risks and roadblocks to improving fairness, as well as some recommendations to mitigate them. How this applications could benefit from the domain of explainability to improve on fairness is outside of the scope of this paper, and is left to future work. \\

\section*{Acknowledgment}
This work was partly supported by VisuAAL ITN H2020 (grant
agreement No. 861091) and the Austrian Research Promotion Agency
(FFG) under the Grant Agreement No. 878730 and 873495
The authors would also like to thank the developers who participated in the interviews.

\bibliographystyle{IEEEtran}
\bibliography{BiasandFairnessinComputerVisionApplicationsoftheCriminalJusticeSystem}
\end{document}